\documentclass[a4paper,twoside]{article}

\usepackage{amssymb}
\usepackage{amstext}
\usepackage{amsmath}
\usepackage{multirow}
\usepackage{times} 
\usepackage{booktabs}
\usepackage{makecell}

\usepackage{natbib}
\let\bibhang\relax
\usepackage{apalike}
\usepackage{algorithm}
\usepackage[hidelinks]{hyperref}
\usepackage{algorithmic}
\usepackage[textsize=tiny]{todonotes}
\usepackage[caption=false]{subfig}
\usepackage[bottom]{footmisc}
\usepackage{url,cleveref} 

\newcommand{\approach}{our approach}

\newcommand{\Approach}{Our approach}

\newcommand{\mll}{<\!\!<} 

\usepackage{SCITEPRESS}     

\begin{document}

\title{Finding Strong Lottery Ticket Networks with Genetic Algorithms}


\author{
\authorname{Philipp Altmann, Julian Schönberger, Maximilian Zorn, and Thomas Gabor}
\affiliation{LMU Munich, Germany}
\email{philipp.altmann@ifi.lmu.de}
} 


\keywords{Evolutionary Algorithm, Neuroevolution, Lottery Ticket Hypothesis, Pruning, Neural Architecture Search}

\abstract{
According to the Strong Lottery Ticket Hypothesis, every sufficiently large neural network with randomly initialized weights contains a sub-network which -- still with its random weights -- already performs as well for a given task as the trained super-network. We present the first approach based on a genetic algorithm to find such \textit{strong lottery ticket} sub-networks without training or otherwise computing any gradient. We show that, for smaller instances of binary classification tasks, our evolutionary approach even produces smaller and better-performing lottery ticket networks than the state-of-the-art approach using gradient information.}

\onecolumn \maketitle \normalsize \setcounter{footnote}{0} \vfill

\section{Introduction}\label{sec:intro}

\begin{figure}[t]
  \centering
  \includegraphics[width=0.9\linewidth]{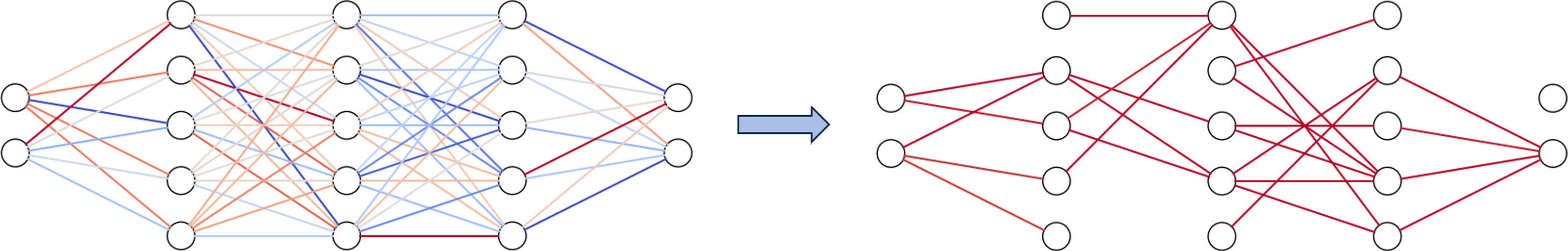}
  \caption{Illustration of a lottery ticket network. Top: Full network graph. Red connections persist in most evolved lottery ticket networks in an example population (blue connections do not). Bottom: Example of an evolved lottery ticket subnetwork with only a fraction of active connections.}
  \label{fig:lottery_ticket}
\end{figure}

A central aspect to the wide success of \textit{artificial neural networks} (ANNs) is that they are usually designed to be \emph{overparametrized} \citep{aggarwal2018neural}. 
That means that they feature more parameters (weights) than are strictly necessary to represent the function they are meant to approximate. 
However, it is also that overparametrization that constructs a solution landscape that is friendly towards relatively simple optimization strategies like stochastic gradient descent \citep{shevchenko2020landscape}, whose application is also enabled by the fact that neural networks are usually differentiable and can thus provide gradient information to the optimization algorithm.
The \textit{Lottery Ticket Hypothesis} \citep{frankle2018lottery} and its variants \citep{ramanujan2020s} have provided a different perspective on the properties of neural networks: Among the randomly initialized weights (before any optimization), some weights have already ``won the lottery'' by being easily trainable. 
Furthermore, in any sufficiently overparametrized network, there already exist --- at the point of random initialization --- certain subnetworks that (when unhinged from the rest of the network) approximates the desired function as accurately as the whole network would \emph{after optimization}. 
Thus, if these subnetworks or \textit{strong lottery tickets} could be found easily, the whole training process of neural networks could be skipped. 
\autoref{fig:lottery_ticket} illustrates a lottery ticket network evolved from a full network with much more active (i.e., non-zero) connections.
\\[2pt]
Finding such subnetworks naturally requires a substantial computational load, as the number of possible combinations of connections to prune from the subnetwork grows exponentially with the network size.
This makes it difficult for a lottery-ticket-based optimization alternative to succeed in practice. In fact, state-of-the-art methods for finding lottery tickets tend to utilize regular training steps of the full network (without changing the weights) to identify more important connections to be kept in the subnetwork.
\\[2pt]
This paper presents a novel approach to finding strong lottery tickets based purely on combinatorial evolutionary optimization without training the weights or utilizing gradient information. To the best of our knowledge, this is the first approach in this direction. 
We summarize our contribution as follows:
\begin{itemize}
\item We show that a basic genetic algorithm (GA) can already produce strong lottery ticket networks.
\item Our approach yields sparser and more accurate networks compared to the gradient-based state-of-the-art in exemplary binary classification tasks.
\item Uncovering scenarios where the utilized GA operations are insufficient, we hope to pave the way for further investigating the applicability of GAs for optimizing neural networks or similar entities.
\end{itemize}

\section{Related Work}\label{sec:rel_work}
The Lottery Ticket Hypothesis has received considerable attention in recent years, and as such, many connections to adjacent fields have been discovered. In this section, we will elaborate on the existing literature and how it relates to our work.

\paragraph{Lottery Ticket Hypothesis}\label{sec:rel_work:lth}
\citet{frankle2018lottery} discovered that a network that was pruned after training and then had its remaining weights reset to their original random-initialized value could then be trained again to achieve a comparable test accuracy to the original network in a similar number of iterations. They called this phenomenon the \textit{Lottery Ticket Hypothesis} (LTH) and the pruned subnetwork a \textit{winning ticket}. They developed an algorithm based on iterative magnitude pruning to find these \textit{winning tickets}. Since then, many approaches have been developed to find these winning tickets: 
\citet{jackson2023finding} use an evolutionary algorithm where they calculate the fitness based on the network density and validation loss in an attempt to deal with the trade-off between the sparsity and the accuracy of the subnetwork. 
Other subsequent work \citep{zhou2019deconstructing, wang2020pruning} extended the LTH by empirically showing that it is possible to find subnetworks that already have better accuracy than random guessing within randomly initialized networks without any training. \citet{zhou2019deconstructing} identify neural network masking as an alternative form of training and introduce the notion of ``supermasks.''
\paragraph{Strong Lottery Ticket Hypothesis}
\citet{ramanujan2020s} built upon this idea and proposed the \textit{Strong Lottery Ticket Hypothesis} (SLTH): 
A sufficiently overparameterized neural network with random initialization contains a subnetwork, the \textit{strong lottery ticket} (SLT), that achieves competitive accuracy (w.r.t. the large, trained network) without any training \citep{malach2020proving}. Additionally, they introduced \emph{edge-popup}, an algorithm for finding strong lottery tickets by approximating the gradient of a so-called pop-up score for every network weight.
These popup scores are then updated via stochastic gradient descent (SGD). A series of theoretical works studied the degree of required overparameterization \citep{malach2020proving, orseau2020logarithmic, pensia2020optimal} and proved that a logarithmic overparameterization is already sufficient \citep{orseau2020logarithmic, pensia2020optimal}. 
On the quest for more efficient methods for finding SLTs, \citet{whitaker2022quantum} proposed three theoretical quantum algorithms that are based on edge-popup, knowledge distillation \citep{hinton2015distilling}, and NK Echo State Networks \citep{whitley2015optimal}. 
Finally, \citet{chen2021peek} introduced an additional type of high-performing subnetwork called ``disguised subnetworks'' that differ from regular SLTSs in the way that they first need to be ``unmasked'' through certain weight transformations. They retrieve these special subnetworks via a two-step algorithm performing sign flips on the weights of pruned networks using Synflow \citep{tanaka2020pruning}. 
\paragraph{Weak Lottery Ticket Hypothesis}
Only a few methods for finding strong lottery tickets have been developed to this point, and most of the empirical work has been focused on the original LTH. They identify so-called \textit{weak lottery tickets} that can achieve competitive accuracies (on the much smaller subnetworks), but only when the subnetworks' weights are re-trained. 
This cycle of training, pruning, and re-training is generally expensive, and the advantages compared to standard training are less obvious. 
In contrast, searching for strong lottery tickets allows one to uncover high-accuracy scoring subnetworks without any (potentially expensive) (re-)training steps. 
Furthermore, its combination with meta-heuristic optimization allows the application to structures of discontinuous functions that would not be learnable via gradient-based approaches. 
In this paper, we propose a method for finding strong lottery tickets that is based purely on genetic algorithms. Existing methods often use heuristics and pseudo-training algorithms that work with some form of gradient descent and usually fix pruning ratios beforehand. In contrast, our approach does not require gradient information, can directly optimize the subnetwork encoding, and does not apply any artificial bound to the maximum number of pruned weights. 
Moreover, genetic algorithms frequently excel at discovering high-quality solutions to NP-hard problems and due to their stochastic nature and global search capabilities are generally well suited for optimizing non-convex objective functions with many local minima, saddle points and plateaus. In the case of the SLTH the optimization landscape is highly complex, with potential non-convexity due to masking, random initialization and the loss function. 
Note that the method of \citet{jackson2023finding}, although very similar to our method, applies the evolutionary algorithm to the original LTH and can thus only find weak lottery tickets that have to be trained.

\paragraph{Extreme Learning Machine}
\citet{huang2006extreme} proposed \textit{Extreme Learning Machine (ELM)}, an approach conceptually similar to the SLTH, where the random parameter values of the hidden layer in a single-hidden-layer neural network are fixed, while the optimal weights of the output layer are calculated using the closed-form solution for linear regression.
In comparison to SLTs, the dense models are less parameter-efficient, do not scale well to deep architectures requiring complex adaptations e.g. based on Autoencoders \citep{kasun2013representational} 
and include the calculation of the matrix inverse which is computationally intensive.

\paragraph{Neural Architecture Search}
There are parallels between neural architecture search (NAS) and searching for lottery tickets since, in both cases, we generate a network of previously unknown structures and untrained (but perhaps selected) weights. \citet{gaier2019weight} investigated the influence of the network architecture compared to the initialization of its parameters when it comes to solving a specific task. They initialized all parameters with a single value sampled from a uniform distribution and concluded they could find architectures that achieved higher-than-random accuracy on the MNIST dataset. \citet{wortsman2019discovering} developed a method that enables continuous adaptation of a network's connection graph and its parameters during training. They showed that the resulting networks outperform manually engineered and random-structured networks. 
Compared to our approach, \citet{gaier2019weight} use a single fixed value for the parameters instead of drawing values from a random distribution. The approach presented by \citet{wortsman2019discovering} is an alternative to finding winning tickets. \citet{ramanujan2020s} then introduced edge-popup inspired by that work, but since their learning of the network structure and its parameterization is inseparable, their approach cannot be used to find a pruning mask for strong lottery tickets.
%
\paragraph{Evolutionary Pruning}
In contrast to NAS or the related field of neuroevolution, both of which typically include evolving the topology of the network, evolutionary pruning solely focuses on pruning the network, i.e., removing connections and possibly whole neurons from the network graph. With such techniques, many networks can be reduced in size without affecting their performance. This branch of research consists of methods that differ in the choice of solution representation (direct encoding or indirect encoding) and the number of objectives. Methods that use direct encoding often work with binary masks that are applied to structures of the network, e.g., single weights or convolution filters \citep{wu2021differential}. Typical multi-objective tasks include, apart from the sparsity goal, also things like accuracy improvement or energy consumption \citep{wang2021evolutionary}. Our approach also works with binary pruning masks and the two objectives, accuracy and sparsity, but to the best of our knowledge, we are the first to apply evolutionary pruning to the setting of the SLTH. 

\paragraph{Other Pruning Methods}
According to \citet{wang2021recent}, besides the classic LTH, which applies static pruning masks on trained networks, and the SLTH, which does not involve any training, there is a third branch of methods that prune at initialization using pre-selected masks \citep{lee2018snip, wang2020picking, tanaka2020pruning}. For example, \citet{lee2018snip} created a pruning mask before training, which zeroed out all structurally unimportant connections, as determined by a new saliency criterion called connection sensitivity. Like our approach, their approach is one-shot since the network only needs to be pruned once, but there is still training involved, and very specific pruning criteria are required to determine good subnetworks.

\section{Method}
\label{sec:method}
In the following, we will discuss the components of the genetic algorithm, including the structure of our solution candidates, the way we determine their fitness and select parents and survivors accordingly, as well as the different genetic operations that guide the evolutional process.

\paragraph{Solution Representation}
\Approach{} generates strong lottery ticket networks via an evolutionary algorithm. We assume that the task that the network is meant to solve is fixed (e.g., given by a classification accuracy function $\mathcal{L}$). We are also given the architecture graph of the full network and the vector of its $n \in \mathbb{N}$ randomly initialized weights $\mathbf{w} = \langle w_0, ..., w_n \rangle$ with $w_i \in \mathbb{R}$ for all $i$. \Approach{} then produces a (genotype) bit mask $\mathbf{b} = \langle b_1, ..., b_n \rangle$ with $b_i \in \{0, 1\}$ for all $i$ so that the (phenotype) masked network $\mathbf{w'} = \langle b_i \cdot w_i \rangle_{i=1,...,n}$ is significantly smaller than the full network w.r.t. non-zero weights, but performs approximately as well as a trained successor of the full network w.r.t. $\mathcal{L}$. Formally, let $\mathbf{w^*}$ be the $n$ weights of the trained full network, then $\mathbf{b}$ should fulfill $\sum_{i=0}^n b_i \mll n$ and $\mathcal{L}(\mathbf{w'}) \approx \mathcal{L}(\mathbf{w^*})$.
Note that we only consider weights in the parameter vector and not any of the potential bias nodes of the network. Yet, although the biases do not get pruned, we still initialize them using our chosen initialization method.

\paragraph{Fitness and Selection}
To drive the evolution of strong lottery tickets, we perform lexicographic evolutionary optimization. We define two objectives: Our primary goal is to find subnetworks that match the accuracies achieved by standard training. Our secondary goal is to retrieve subnetworks that are as sparse as possible without having a negative impact on the accuracy. This multi-objective approach allows us to prune subnetworks by a considerable margin even after very high accuracies have already been achieved. The evaluation of the individuals happens in two places in our evolutionary pipeline: For parent selection (i.e., selecting the individuals for recombination), we only consider the accuracy goal, whereas for survivor selection (i.e., selecting the individuals for the next generation), we also consider the sparsity goal. This accounts for the fact that recombination is the main contributor to better-performing individuals throughout the evolution. Focussing solely on the accuracy goal for parent selection leads to an effective prioritization. The fitness corresponds to the measured accuracy on the train dataset, and the individuals are ranked accordingly. Even though we also consider the sparsity for survivor selection, accuracy is still the main determinant, i.e., for survivor selection, we prefer individuals with a higher sparsity value \textit{within} groups of individuals with the same accuracy.
\\[2pt]
We use (elitist) cut-off selection for survivor selection\footnote{We also tried other selection methods, like roulette or random walk selection, but we found that the choice of selection method had no significant impact.}.
This method selects the top $k$ individuals of the current population and transfers them to the next generation's population. 
In our case, $k = N$ where $N$ is the original population size; since none of our genetic operators are in place, the population typically grows beyond its original size $N$ in between generations and needs to be reduced for the next generation. For parent selection, any individual may be chosen as a first parent with a chance $\texttt{rec\_rate} \in [0,1]$ and matched with a second parent chosen randomly from the top $l$ individuals in the current population where $l = N \cdot \texttt{par\_rate}$ is defined via a hyperparameter $\texttt{par\_rate}$.

\vspace{-2pt} 
\paragraph{Genetic Operators}\label{subsec:gen_and_var}
We implement two steps to generate our initial population: First, the individuals are generated randomly, i.e., each bit has an equal likelihood of being chosen at any given point in the pruning mask. Second, from the randomly generated individuals, we discard those that do not reach a certain \textit{accuracy bound}. In our implementation, we choose to use an adaptive bound that can decrease dynamically if too few individuals match the boundary value, following the shape of a pre-defined exponential function, to reduce the effects that random sampling has on runtime. Using the adaptive accuracy bound allows for a higher initial bound and proved to have a positive influence on the final accuracies. For the following, we refer to the configuration that performs only the first step as \textbf{GA}, 
and the configuration that uses an adaptive accuracy bound (i.e., the first and the second step) is named \textbf{GA (adaptive AB)}
\footnote{All required implementations are available at \url{https://github.com/julianscher/SLTN-GA}.}.
\\[2pt]
We perform single-point mutation, randomly selecting individuals from the current population at a chance $\texttt{mut\_rate}$ and generating a mutant via one random bit flip. 
For recombination, we use random crossover on two parents. Note that mutants and children are always added to the population and never directly replace their parents.
Finally, to further increase the diversity in the population, we add $m$ freshly generated individuals to the population in each generation. The value of $m = N \cdot \texttt{mig\_rate}$ is given by the hyperparameter $\texttt{mig\_rate}$.


\begin{figure*}\centering
  \subfloat[Moons\label{fig:moons}]{\includegraphics[width=0.22\linewidth]{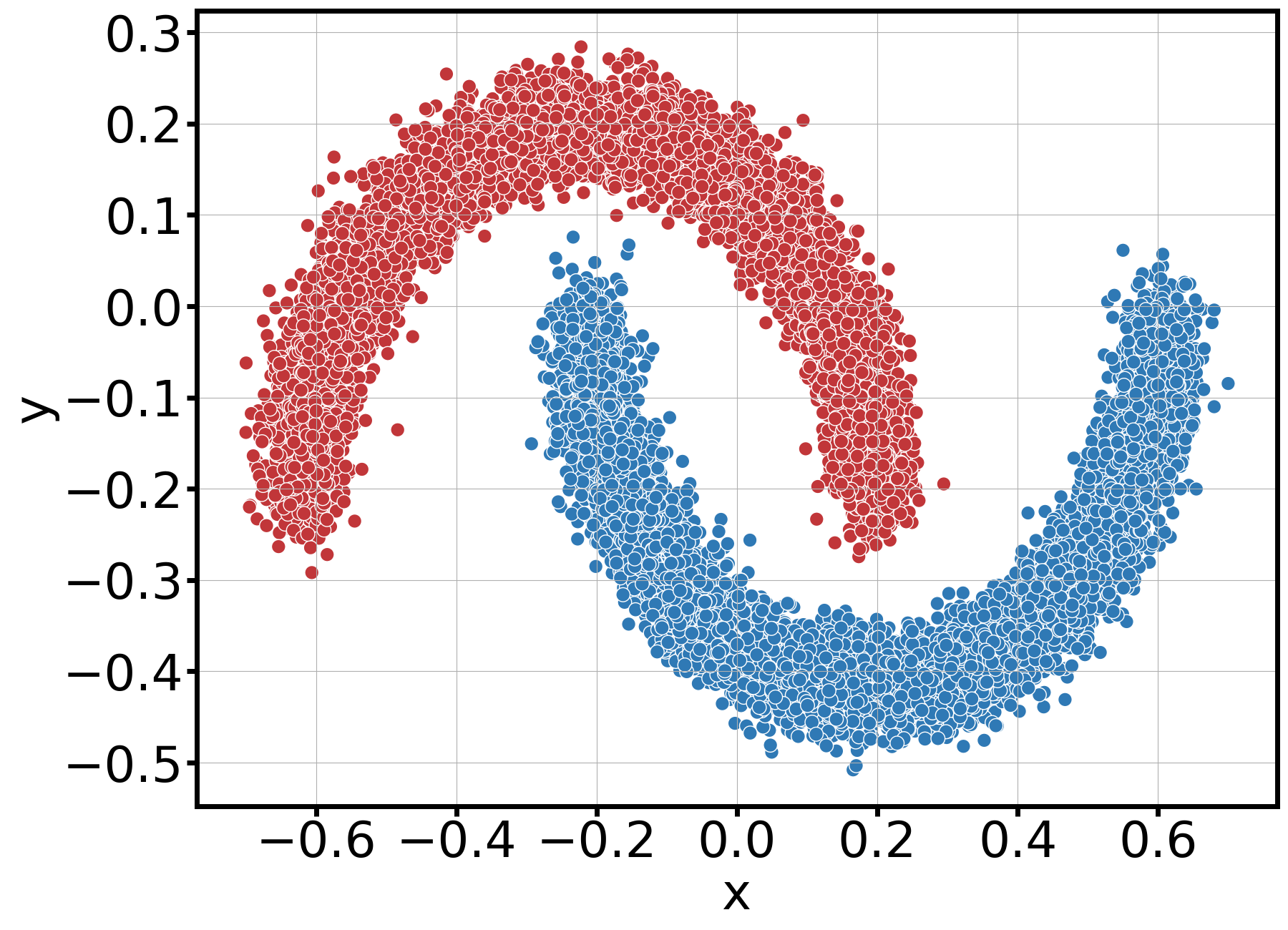}}\hspace{16pt}
  \subfloat[Circles\label{fig:circles}]{\includegraphics[width=0.22\linewidth]{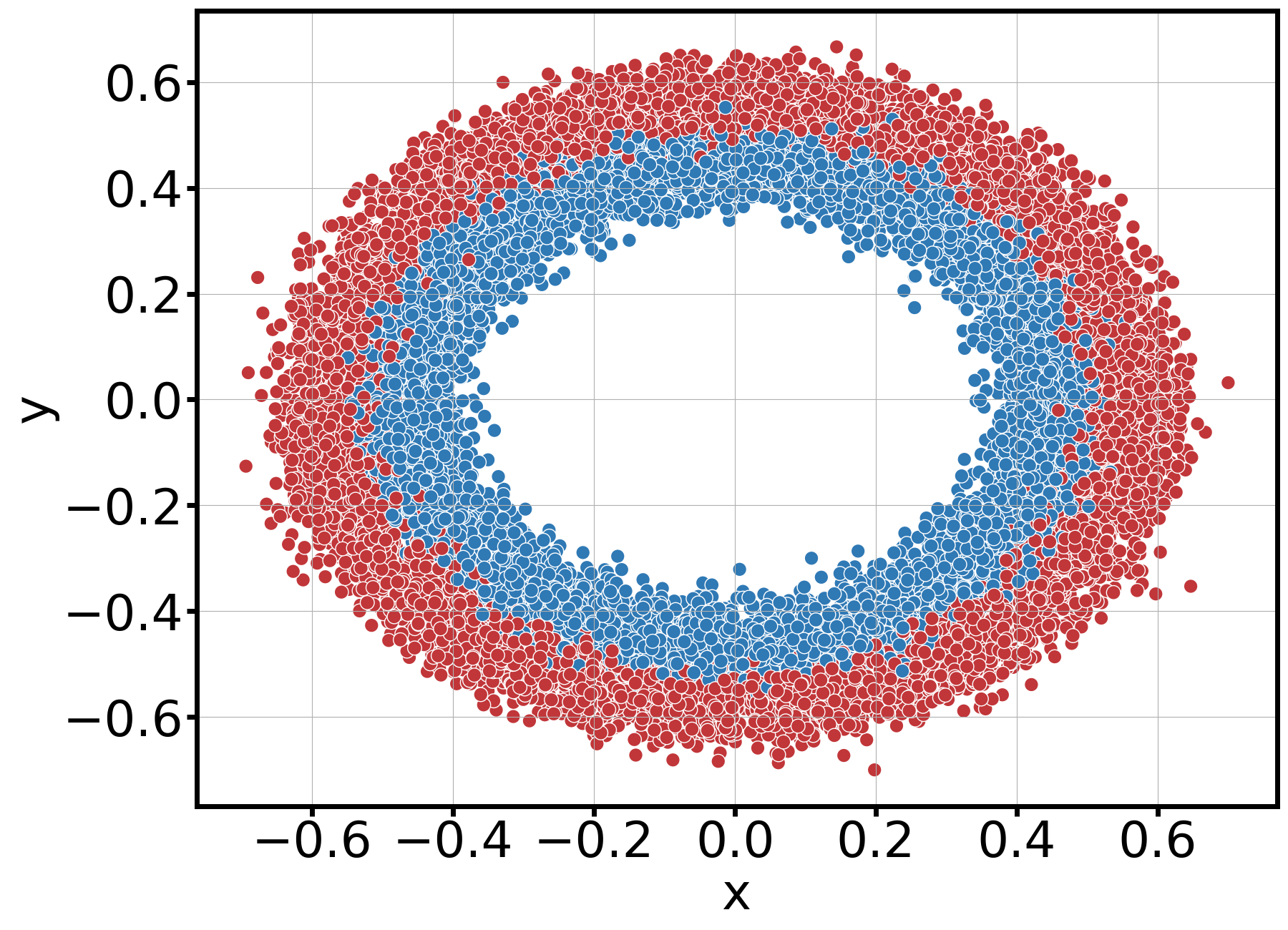}}\hspace{16pt}
  \subfloat[Network architectures\label{tab:architectures}]{
    \raisebox{3.8em}[0pt][0pt]{\small \begin{tabular}{ccc} \toprule 
    Architecture & \texttt{moons}, \texttt{circles} & \texttt{digits} \\ \midrule
    \small A & \footnotesize [2, 20, 2] & \footnotesize [64, 20, 10] \\
    \small B & \footnotesize [2, 75, 2] & \footnotesize [64, 75, 10] \\
    \small C & \footnotesize [2, 100, 2] & \footnotesize [64, 100, 10] \\
    \small D & \footnotesize [2, 50, 50, 2] & \footnotesize [64, 50, 50, 10] \\ \bottomrule
  \end{tabular}}}
  \caption{Overview of our datasets and network architectures we use on them: The \texttt{moons}~(\subref{fig:moons}) test dataset consisting of 16000 2d-datapoints, normalized on the interval $[-0.7, 0.7]$, and the \texttt{circles}~(\subref{fig:circles}) test dataset consisting of two different-sized rings with 16000 2d-datapoints from \texttt{scikit-learn} \citep{pedregosa2011scikit}, and the network architectures~(\subref{tab:architectures}) with a single-lettered identification code. The bracket notation describes the number of neurons in the different network layers. The first number corresponds to the number of input neurons. The last is the number of output neurons. 
    }
\end{figure*}

\section{Experimental Setup}\label{sec:exp}
To evaluate the capabilities of the previously discussed genetic algorithm in finding SLTs, we apply it to multiple datasets and different network architectures. The performance is then compared to the state-of-the-art approach. We conclude with an analysis of the implications of having more than two classes.

\paragraph{Hyperparameters}
For the following experiments, our GA works with a fixed population size of $N = 100$ individuals. Additionally, we use fixed rates for parent selection, recombination, mutation, and migration: For recombination, we use a $\texttt{rec\_rate} = 0.3$, which implies that around $30\%$ of individuals from the whole population are chosen to become a first parent. Due to $\texttt{par\_rate} = 0.3$, then the recombination mate of any first parent is randomly chosen from the top $30\%$ of the population. We choose $\texttt{mute\_rate} = 0.1$ so that approximately $10\%$ of the individuals generate a mutant to be added to the population. That is a fairly high value, but we intend to generate highly explorative runs. For the same reason, we set the $\texttt{mig\_rate} = 0.1$ so that around $10\%$ of the interim population before survivor selection is made up of freshly generated individuals. Table \ref{tab:GA_hyperparameters} summarizes the chosen hyperparameter values. Our GA terminates if the population evolved for at least 100 generations with no accuracy improvement in the last $50$. When using GA (adaptive AB), we restrict the evolution to take at maximum 200 generations, due to us observing that from that point forward the accuracy improvement usually is only marginally and does not justify the additional runtime costs. 
We did not perform any explicit hyperparameter search for determining optimal values, but based our decisions on observations made throughout the implementation phase. This allows us to reason about the general performance of the GA, which can be expected even with potentially suboptimal hyperparameter values.

\begin{table}[h]
    \centering
    \small
    \begin{tabular}{cc}
        \toprule
         Hyperparameter &  Value \\
         \midrule
         $\texttt{pop\_size}$ $N$ & 100 \\
         $\texttt{rec\_rate}$ & 0.3 \\
         $\texttt{par\_rate}$ & 0.3 \\
         $\texttt{mut\_rate}$ & 0.1 \\
         $\texttt{mig\_rate}$ & 0.1 \\
         \bottomrule
    \end{tabular}
    \caption{The used hyperparameters for our GA evaluation.}
    \label{tab:GA_hyperparameters}
\end{table}

\vspace{-10pt} 
\paragraph{Datasets}
Our experiments are built on three datasets with varying complexities. 
We chose classification tasks as they can be easily interpreted and come with a clear and tried evaluation metric. 
The two-dimensional \texttt{moons} dataset with only two classes, depicted in Figure~\ref{fig:moons}, consists of two moon-shaped point clusters with little to no overlap. A 2-layered network with only $6$ hidden cells trained via backpropagation already achieves approximately $100\%$ accuracy in some runs. 
In contrast to this rather simple dataset, we selected the \texttt{circles} dataset, presented in Figure~\ref{fig:circles} as a more challenging 2d binary classification problem. The two classes are arranged as two Gaussian-shaped rings, where the bigger ring surrounds the smaller ring. The transition is immediate, and there are many overlapping points, which is a challenging task even for the trained dense network. We generate $66000$ random data points for both datasets and add Gaussian noise with $\sigma = 0.07$. As a third dataset, we use the \texttt{digits} dataset, which consists of $1797$ images with size $8 \times 8$ pixels each and class labels $\{0, ..., 9\}$.  
We split the datasets into a training and a test dataset, using $25\%$ of the data points for testing. Additionally, we perform min-max normalization on the \texttt{moons} and the \texttt{digits} datasets to mitigate potentially negative scaling effects for the networks, which can arise from non-Gaussian distributions.

\paragraph{Network Architectures}
We only use classical feed-forward ANNs with ReLU activation for the neurons in the input and hidden layers. Since for the GA we are primarily interested in the final accuracies and not the class probabilities we do not use a softmax activation function, but instead, calculate the accuracies directly using the class of the highest valued network output. In order to get a better intuition about the GA's behavior across different model sizes, we test $4$ network architectures as listed in Table~\ref{tab:architectures} in our experiments. For simplicity, we only denote the analyzed network architectures by ``A'', ``B'', ``C'', and ``D'' in the later plots. 
Our studies showed that the choice of the network parameter initialization method greatly impacts the achieved final accuracies. We sample the network weights from a uniform distribution over the interval $[-1, 1]$ for all our GA experiments. This method proved to yield the best overall results on the considered datasets. Additionally, there already exist proofs for the existence of SLTs based on uniform parameter initializations \citep{malach2020proving, pensia2020optimal}. Although most of the work on the SLTH works with zeroed-out biases, we experienced a significant performance boost when we initialized the biases by sampling from the same uniform distribution. 

\paragraph{Baselines}


Finally, since, by definition of strong lottery tickets, we are particularly interested in the comparative performance of a network that was trained using a gradient-based method, we use backpropagation as a baseline. To compare against a sophisticated implementation of a trainable feed-forward network, we used the \textit{MLPClassifier} module from \textit{scikit-learn} and performed hyperparameter tuning on all $4$ architectures using their \textit{RandomizedSearchCV} function. We employ random search because of its computational efficiency in exploring large parameter spaces with a limited computation budget. The chosen parameter ranges were selected based on prior knowledge and preliminary experiments. Specifically, the tuned hyperparameters include solvers, learning rates, batch sizes, momentum, alphas (for l2 regularization) and epsilon values (for numerical stability). An overview of the resulting values is provided by Table \ref{tab:backprop_hyperparameters}. The search and the subsequent training lasted $1000$ epochs to ensure convergence. Our studies compare the mean accuracies of the backpropagation trained networks from Table \ref{tab:architectures} on the test datasets. 
\begin{table}[h]
\centering
\resizebox{\columnwidth}{!}{%
\begin{tabular}{cccccccc}
\toprule
Dataset & Solver & Learning Rate & Learning Rate Init & Epsilon & Batch Size & Alpha & Momentum \\
\midrule
\multirow{4}{*}{moons} 
& adam & constant & 0.021544 & 4.64e-09 & 128 & 0.0001 & - \\
& adam & constant & 0.001 & 4.64e-09 & 64 & 0.000215 & - \\
& adam & constant & 0.001 & 4.64e-09 & 64 & 0.000215 & - \\
& adam & constant & 0.001 & 4.64e-09 & 64 & 0.000215 & - \\
\midrule
\multirow{4}{*}{circles} 
& sgd & adaptive & 0.1 & - & 64 & 0.046416 & 0.0 \\
& sgd & adaptive & 0.004642 & - & 128 & 0.046416 & 0.5, nesterov \\
& adam & constant & 0.001 & 4.64e-09 & 64 & 0.000215 & - \\
& sgd & adaptive & 0.1 & - & 128 & 0.046416 & 0.0, nesterov \\
\bottomrule
\end{tabular}
}
\caption{Listing of the determined backpropagation hyperparameters for the MLPClassifier model from \textit{scikit-learn} using random search.}
\label{tab:backprop_hyperparameters}
\end{table}


\begin{figure*}[t]\centering
\subfloat[\texttt{moons}, $R = 50$\label{fig:GA_moons}]{\includegraphics[width=0.28\linewidth]{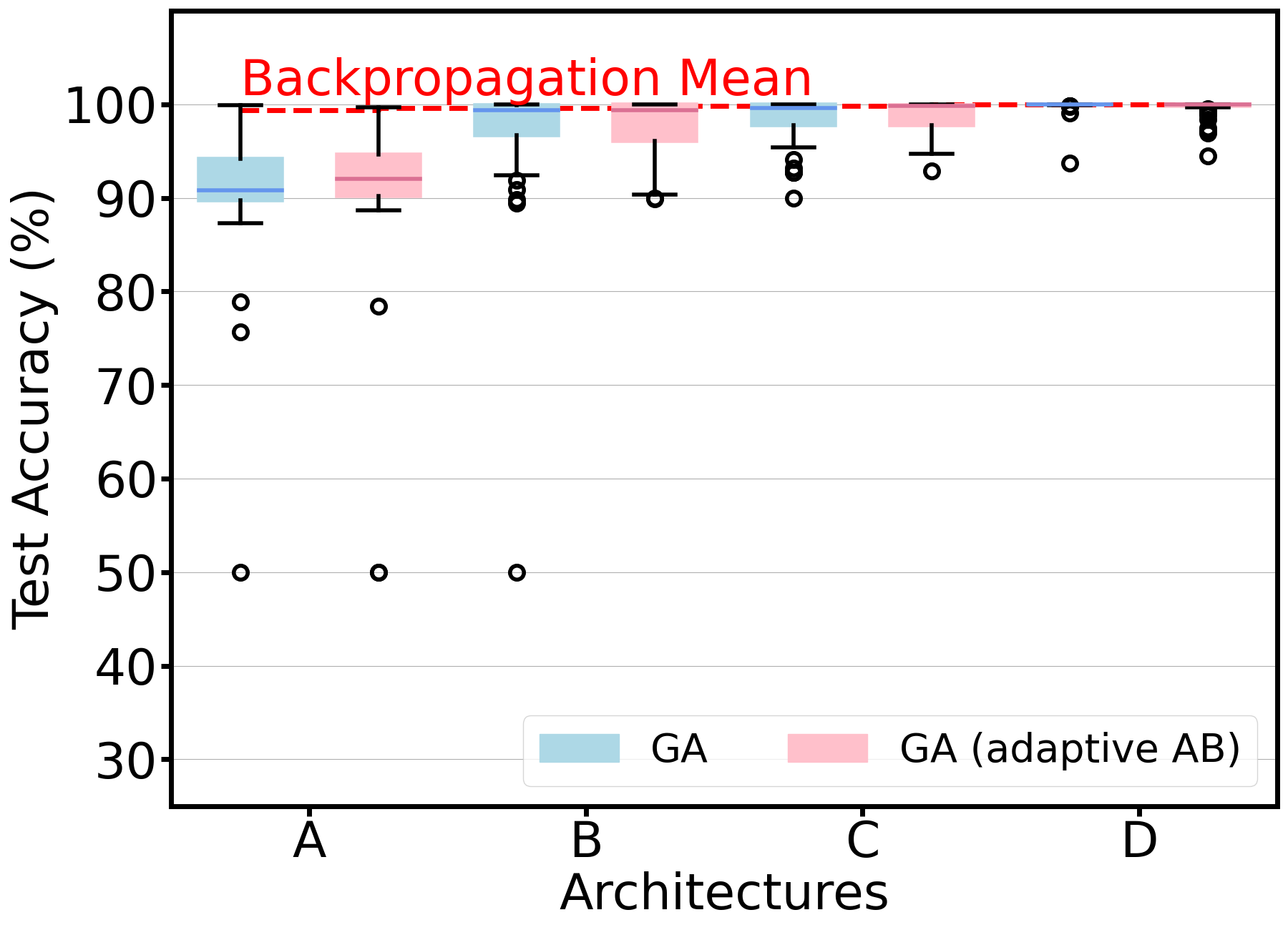}}
\subfloat[\texttt{circles}, $R = 50$ \label{fig:GA_circles}]{\includegraphics[width=0.28\textwidth]{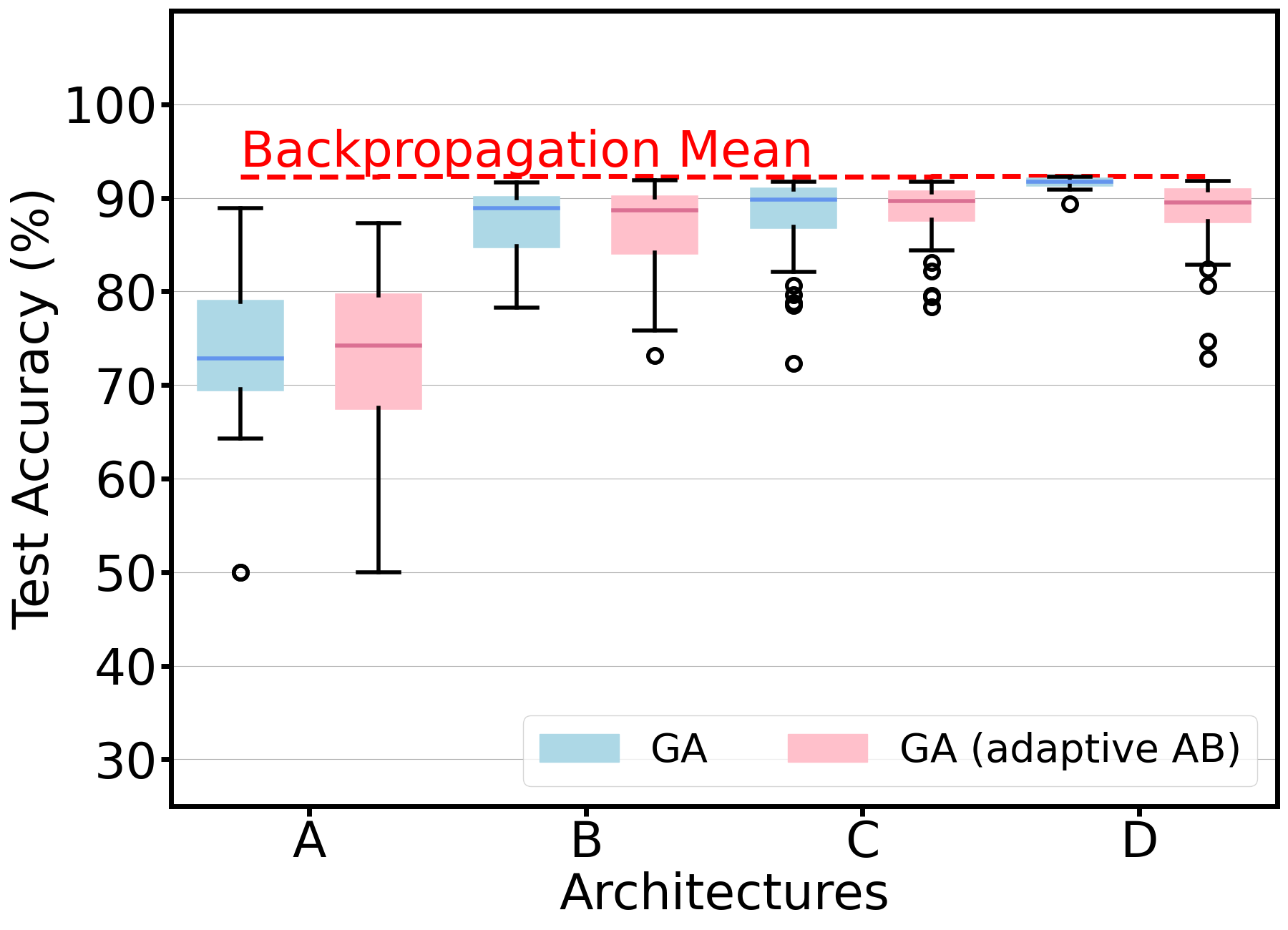}}
\subfloat[Reached mean accuracies $\pm$ standard deviation\label{tab:GA_accuracies}]{
\raisebox{48pt}[0pt][0pt]{
\resizebox{0.4\linewidth}{!}{
\tiny
\begin{tabular}{ccccc}\toprule
 & Arch. & GA & GA (adaptive AB) & Backprop.\\
 \midrule
 \parbox[t]{2mm}{\multirow{4}{*}{\rotatebox[origin=c]{90}{\texttt{moons}}}} 
 & A & 90.7\% $\pm$ 7.2 & 90.9\% $\pm$ 9.1 & 99.4\% $\pm$ 2.4 \\
 & B & 96.9\% $\pm$ 7.3 & 97.5\% $\pm$ 3.3 & 99.6\% $\pm$ 1.9 \\
 & C & 98.4\% $\pm$ 2.5 & 98.9\% $\pm$ 1.7 & 99.8\% $\pm$ 1.4 \\
 & D & 99.8\% $\pm$ 0.9 & 99.6\% $\pm$ 1.0 & 99.9\% $\pm$ 0.0 \\
 \midrule
\parbox[t]{2mm}{\multirow{4}{*}{\rotatebox[origin=c]{90}{\texttt{circles}}}} 
 & A & 73.9\% $\pm$ 8.1 & 73.7\% $\pm$ 8.6 & 92.3\% $\pm$ 0.1 \\
 & B & 87.3\% $\pm$ 3.6 & 86.8\% $\pm$ 4.4 & 92.4\% $\pm$ 0.0  \\
 & C & 88.0\% $\pm$ 4.3 & 88.5\% $\pm$ 3.2 & 92.3\% $\pm$ 0.0  \\
 & D & 91.6\% $\pm$ 0.5 & 88.3\% $\pm$ 4.0 & 92.4\% $\pm$ 0.0  \\
\bottomrule
\end{tabular}
}
}
}
\caption{Overview of the performance of the GA in the moons~\subref{fig:GA_moons} and circles~\subref{fig:GA_circles} datasets. The blue boxes contain different runs for every architecture using the default GA configuration. The pink boxes contain the results of $R$ runs for the GA configuration that uses the adaptive accuracy bound with initial threshold value $0.85$. For comparison, we added the mean accuracies that were achieved with the trained networks using backpropagation. \subref{tab:GA_accuracies} summarizes the achieved accuracies.}
\label{fig:GA_perform}
\end{figure*}

\section{Experimental Results}

\subsection{GA Performance Analysis}\label{subsec:GA_perf}
\begin{figure*}[t]\centering
    \subfloat[Accuracy\label{fig:acc_dev}]{\includegraphics[width=0.42\textwidth]{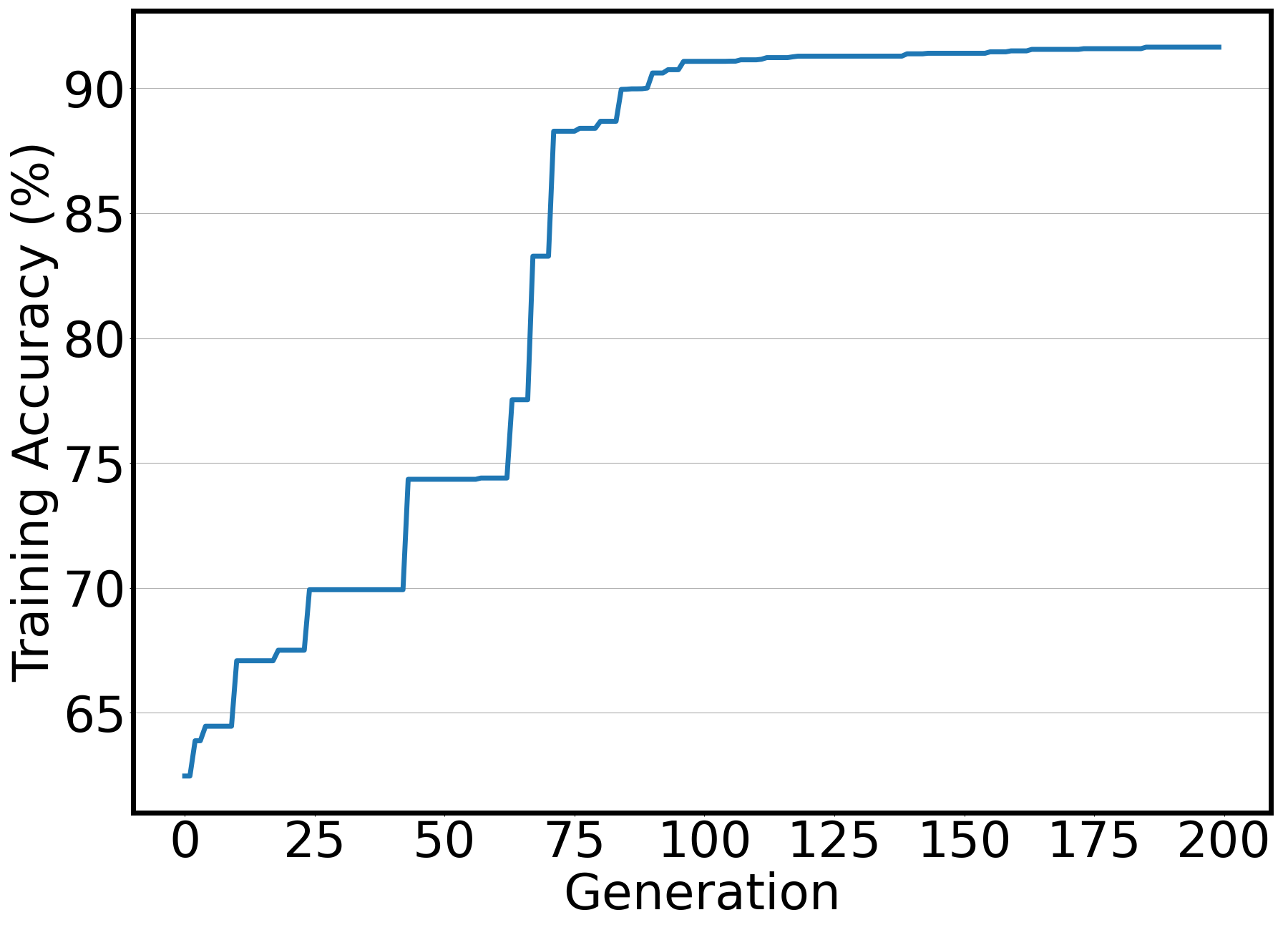}}\hspace{0.05\textwidth}
    \subfloat[Sparsity\label{fig:spars_dev}]{\includegraphics[width=0.42\textwidth]{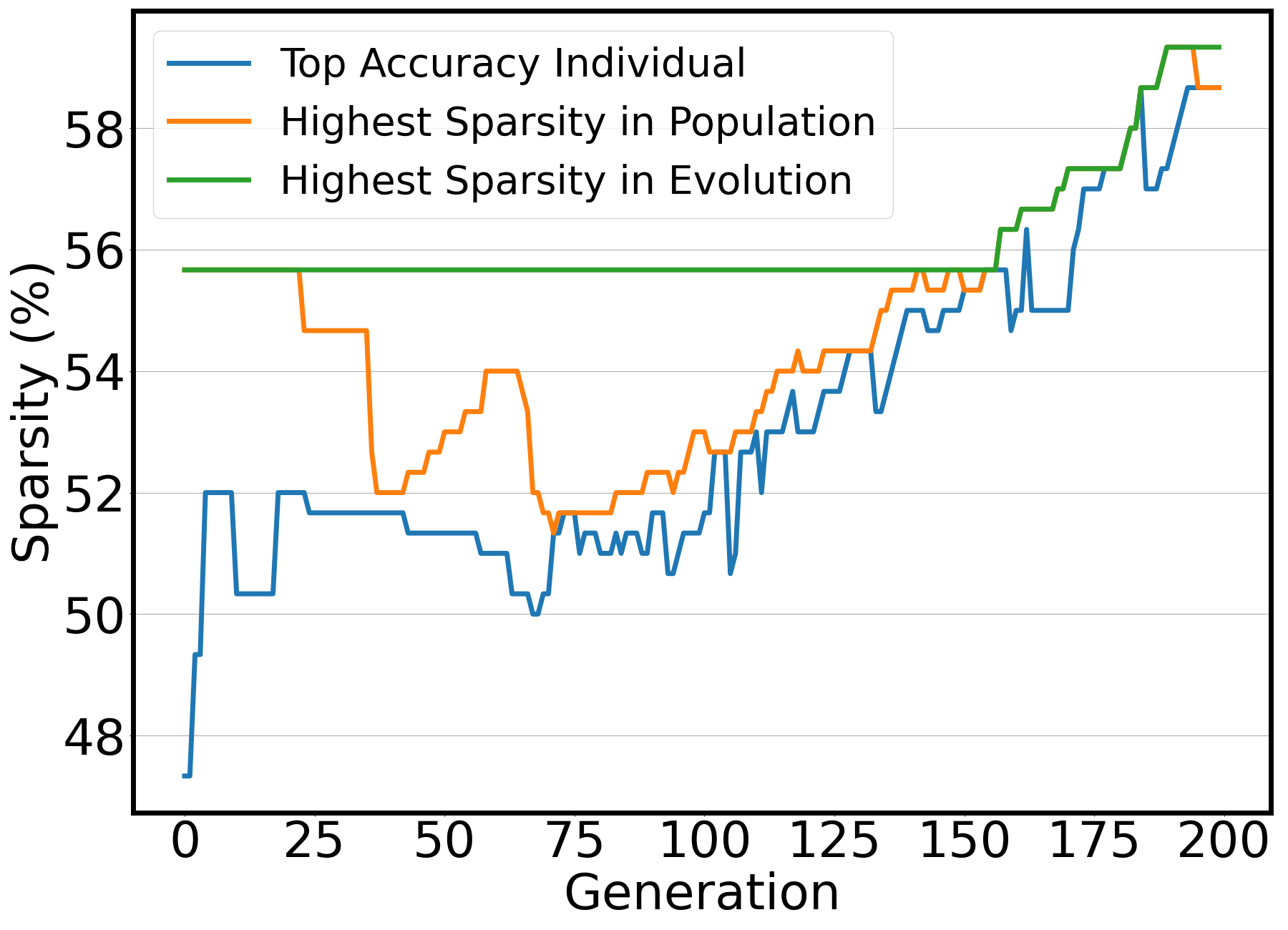}}
    \caption{Optimization progress of one well-performing run using ``GA (adaptive BA)'' on network architecture B $=[2, 75, 2]$ in the \texttt{circles} dataset with regard to the accuracy \subref{fig:acc_dev}, and the sparsity \subref{fig:spars_dev}. The blue line shows the sparsity of the fittest individual in the current population. The orange line displays the top sparsity in the current population, and the green line represents the current best sparsity found in all previous generations.}
    \label{fig:GA_example_run}
\end{figure*}

As mentioned previously, we use 4 different network architectures in our experiments (cf. Table~\ref{tab:architectures}). The general intuition would be that networks with higher parameter counts are more likely to contain parameters with lucky initializations, leading to higher-scoring subnetworks. Additionally, we are interested in whether the usage of an accuracy bound for the generation of the initial population has a noticeable impact on the subsequent evolution. 
\\[2pt]
The results for the \texttt{moons} dataset are shown in Fig.~ \ref{fig:GA_moons}. We observe that the GA is able to achieve very high final accuracies, reaching almost 100\% mean accuracy for network D. Examining the distribution of the different GA runs for the various network architectures, there exists a clear connection between the number of network parameters and the performance. Whereas, for the smallest network A with only 80 parameters, the mean difference to backpropagation is around 9\%. The difference diminishes continuously with increasing parameter count. For networks C and D, the mean approximately matches that of backpropagation, and for network D, there remains only little variance between the runs. The difference in performance between the different GA configurations is less prominent. In general, the mean for the runs using an accuracy bound is a little higher than those that did not use it, but for increasing network sizes, this effect plays less of a role.
\\[2pt]
The results on the \texttt{circles} dataset, illustrated in Fig.~\ref{fig:GA_circles}, mostly support these findings. Considering the mean performance of backpropagation, it becomes clear that the \texttt{circles} dataset has higher complexity than the \texttt{moons} dataset. The GA, again, scores the lowest accuracies on network architecture A but reaches higher final accuracies on the larger networks. The highest mean accuracy of $91.6\%$ is achieved on network D, but this time without using an accuracy bound. Also, there seems to be a certain minimum threshold for the parameter count before which the final accuracies are noticeably lower, but increasing the network size has less of an effect after exceeding it. Still, we can say that there exist situations where the GA is able to score very similar accuracies to backpropagation.  
\\[2pt]
To get an impression of the typical behavior of the GA regarding the development of our accuracy and sparsity objectives, we selected one high-performing example run from the runs on the \texttt{circles} dataset; that run was performed on network architecture B using the GA configuration with an adaptive accuracy bound. In Fig.~\ref{fig:acc_dev}, we see that the individual with the highest fitness in the initial population had less than $65\%$ accuracy. This accuracy is then successively improved in the first 100 generations, taking a set of big leaps until the final accuracy reaches a plateau at around $91\%$ accuracy. This clearly shows the optimization capabilities of the genetic algorithm. For the next 100 generations, until the generation threshold for GA (adaptive AB) is reached, only minor improvements are made. Meanwhile, Fig.~\ref{fig:spars_dev} shows how the sparsity develops over the course of the evolution. Typical behavior is that the sparsity decreases in the first half of the generations since we prioritize achieving our accuracy goal, and only when the improvement of the accuracy slows down does the optimization of the sparsity really start to show. GA really started to improve on the sparsity objective. That is because, at that point, the population is very homogeneous, and there are many individuals with the same accuracy. In this run, the GA achieved an additional improvement of around $10\%$ in sparsity compared to the top individual in the initial population.

\paragraph{Scalability}
Calculating the fitness of the individuals in the population is the decisive factor on runtime complexity with $\mathcal{O}(g*N*(d*l*b^{2}))$ multiplications for an evolution with $g$ generations, a population of size $N$, $d$ dataset samples and a worst case network architecture with $l*b^{2}$ parameters (i.e., length of the bit-vector). Typically $N < g$ and $(l*b^{2}) \ll d$. In practice, the effect of $g*N$ on the runtime can be reduced by efficient parallelization. A compressed version of the subnetwork encoding reduces the complexity for the other GA operations.

\subsection{Edge-Popup \& Weight Initialization}\label{subsec:Edge_popup_perf}
In the previous subsection, we saw that the GA performs well on the given binary classification tasks, achieving accuracies that are very close to or even match the accuracies obtained by training via backpropagation, given a sufficient network architecture is chosen. To get an idea of how well the GA performs in comparison to other methods that search for SLTs in a randomly initialized neural network, we repeat our previous experimental setup using the well-known edge-popup algorithm \citep{ramanujan2020s}. Edge-popup assigns a score to each weight of the neural network and constructs subnetworks by only choosing the top $k\%$ scoring edges in each layer for the forward pass. The scores are updated in the backward pass by using the straight-through gradient estimator \citep{bengio2013estimating}. Once pruned, edges can re-appear in a subnetwork since the edges' contribution to the loss is continuously re-evaluated when approximating the gradients. The parameter $k$ in the forward pass denotes a fixed value, which is also called the pruning rate. Therefore, a pruning rate of 60\% corresponds to a subnetwork where $(1 - k) = 40\%$ of weights are pruned. Note that the sparsity metric we use in our work works the other way around. A subnetwork with a sparsity of 60\% means 60\% of weights are pruned.
\begin{figure*}[t]\centering
    \subfloat[\texttt{moons, $R = 25$}\label{fig:edge_moons}]{\includegraphics[width=0.42\textwidth]{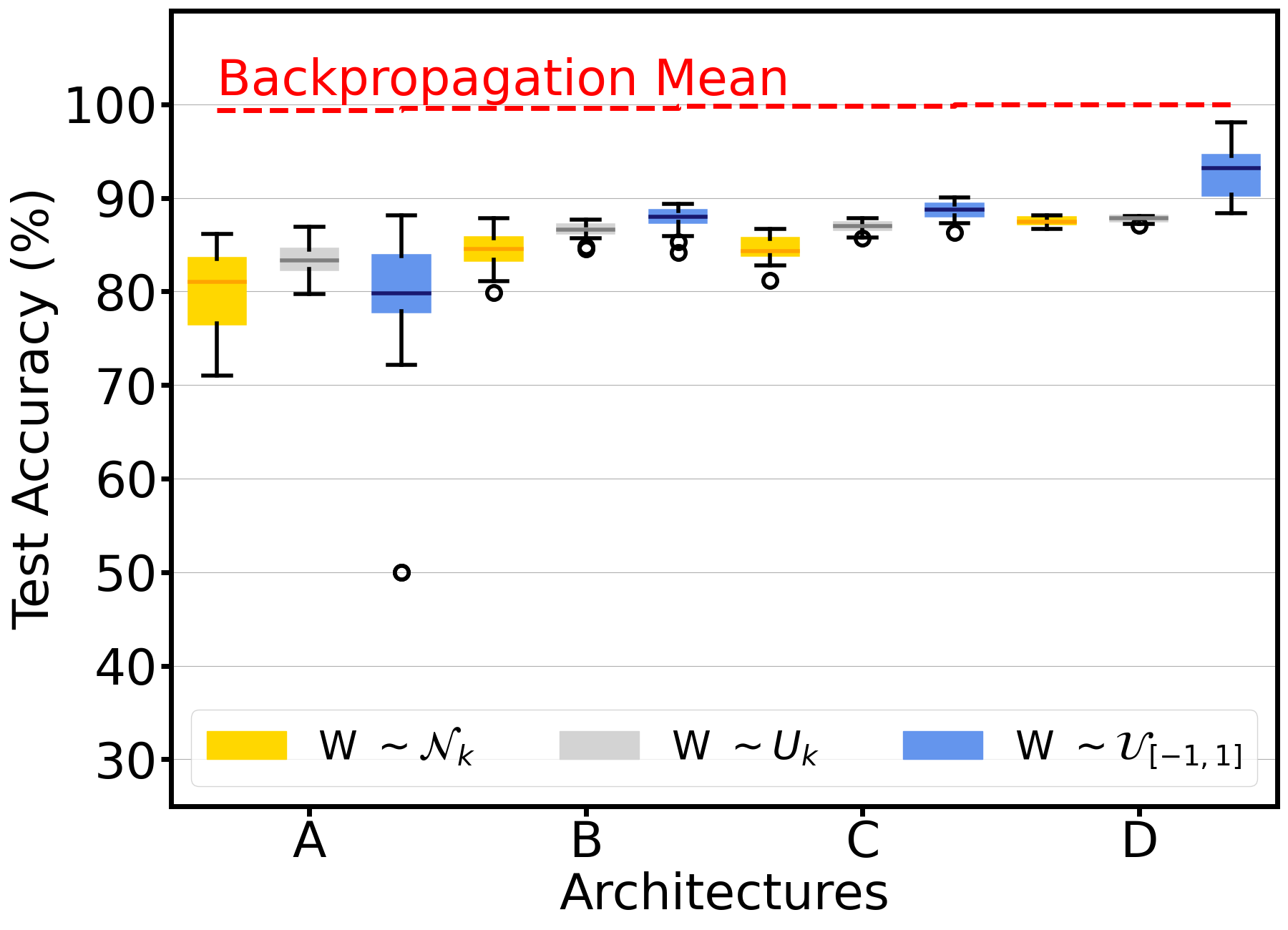}}\hspace{0.05\textwidth}
    \subfloat[\texttt{circles, $R = 25$}\label{fig:edge_circles}]{\includegraphics[width=0.42\textwidth]{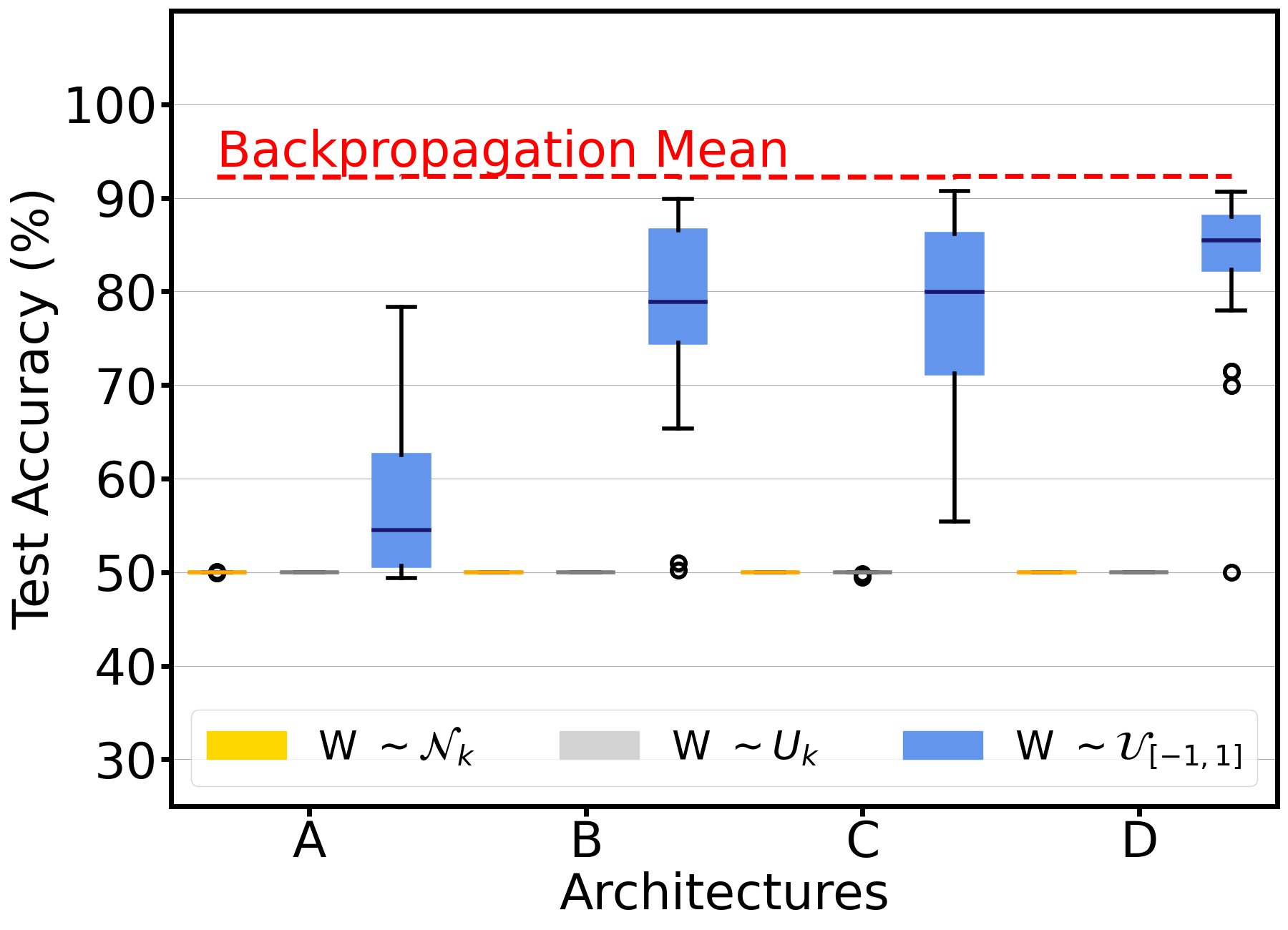}}
    \caption{Illustration of the performance of edge-popup on shown datasets using the different color-coded initializations with $R$ runs each. The backpropagation mean accuracies on the respective architectures (dashed line) are provided for comparison.}
    \label{fig:edge_perform}
\end{figure*}
\\[2pt]
We use the default settings from the authors and train for a total of $100$ epochs. Every configuration is evaluated on $25$ random seeds. The authors found two initialization methods that worked particularly well for their experiments: initializing the network parameters from a Kaiming normal distribution (also known as He initialization \citep{he2015delving}), which (following the notation of \citet{ramanujan2020s}) we refer to as ``Weights $\sim \mathcal{N}_{k}$'', or sampling from a signed Kaiming constant distribution, which we refer to as ``Weights $\sim U_{k}$''. Thus, in addition to using our initialization method, which we indicate as ``Weights $\sim \mathcal{U}_{[-1, 1]}$'', we also consider runs where the networks are initialized using both of their methods. Note that we use the scaled versions of these methods, where the standard deviation is scaled by $\sqrt{1/k}$. For the exact definitions of these methods, refer to \citet{ramanujan2020s}. As with our GA, we also sample the biases from the uniform distribution when using our parameter initialization method with edge-popup. 
\\[2pt]
Due to the considerable performance difference between alternative parameter initializations for our GA, which is also supported by the findings of \citet{ramanujan2020s}, we start with an ablation study to determine the highest accuracy achieving initialization technique for edge-popup before proceeding with the actual comparison studies. Fig.~\ref{fig:edge_moons} shows the results of the different runs edge-popup on the \texttt{moons} dataset. The trend that the larger the network, the higher the final accuracies on the \texttt{moons} dataset typically are, seems to apply here as well. It is also noticeable that the runs using our parameter initialization method (apart from network A) generally outperformed the other run-throughs. In the case of network D, it did so by quite a significant margin ($\approx 5\%$ mean difference). Nevertheless, in none of the settings, edge-popup's mean accuracy comes close to the performance of backpropagation. 
The same holds for the \texttt{circles} experiment, as shown in Fig.~\ref{fig:edge_circles}, except here, the Kaiming normal and signed Kaiming constant distributions proved to be completely insufficient. There is no run where the classification accuracy is better than random, i.e., the predicted class label is correct in only $50\%$ of cases. Considering these results, one might assume that this is an algorithmic issue, but since edge-popup performs well with our initialization method, the issue has to be the Kaiming normal or Kaiming singed constant distributions. A potential reason might be the Gaussian nature of the rings, which has a distorting effect on the methods. Finding the exact cause remains subject to future work.
In summary, it seems that at least for the \texttt{moons} and \texttt{circles} datasets, edge-popup benefits from using uniform initialization. Going forward, we, therefore, decided to sample the parameters for both the GA and edge-popup from the same distribution.
\\[2pt]
\begin{figure*}[t]\centering
  \subfloat[\texttt{moons}\label{fig:edge_moons_adapted}]{\includegraphics[width=0.42\textwidth]{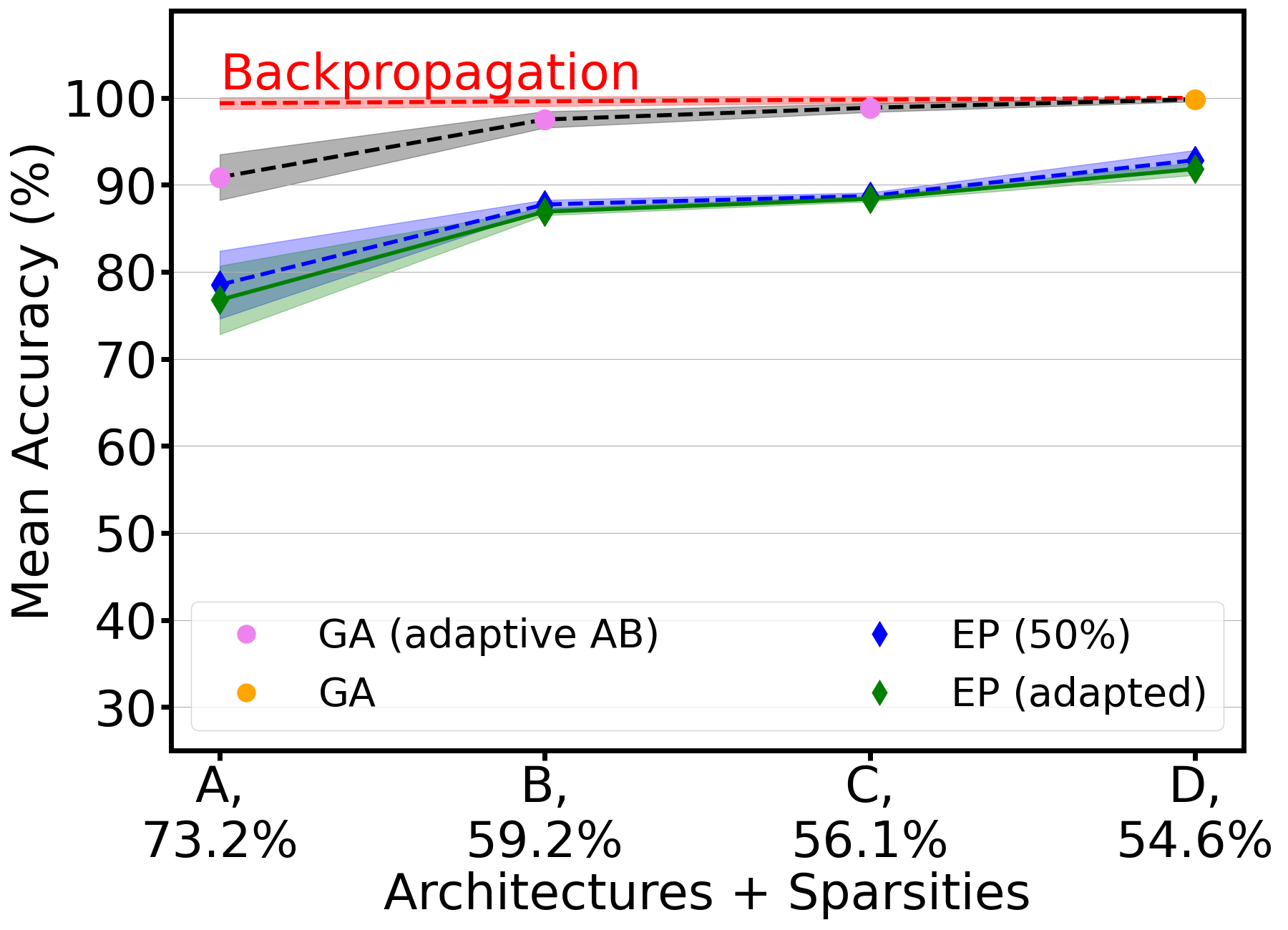}}
  \hspace{0.05\textwidth}
  \subfloat[\texttt{circles}\label{fig:edge_circles_adapted}]{\includegraphics[width=0.42\textwidth]{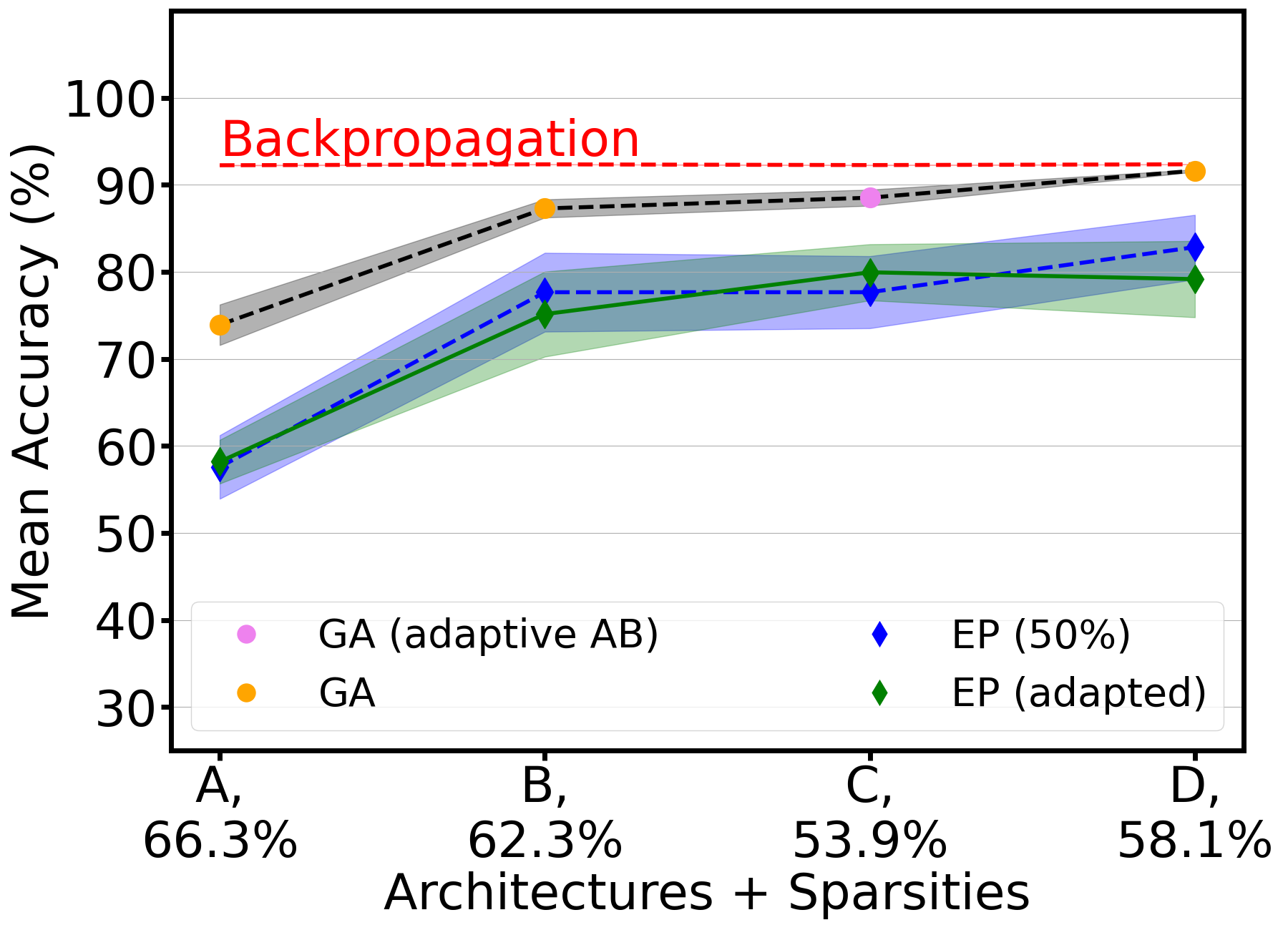}}\\
  \subfloat[Statistical analysis\label{tab:statistics}]{\centering
    \small\begin{tabular}{cccccccc}
        \toprule
        Dataset & Reference & Target & Coef. & Std. Err. & $z$ & $P > |z|$ & 95\%-Conf. \\
        \midrule
        \multirow{4}{*}{moons} & GA & GA (adaptive AB) & 0.042 & 0.083 & 0.507 & 0.612 & [-0.121, 0.205] \\
         & EP (50\%) & EP (adapted) & -0.133 & 0.095 & -1.405 & 0.160 & [-0.320, 0.053] \\
         & GA & EP (50\%) & -1.185 & 0.081 & -14.697 & 0.000 & [-1.343, -1.027] \\
         & GA & Backpropagation & 0.661 & 0.087 & 7.609 & 0.000 & [0.491, 0.831] \\
        \midrule
        \multirow{4}{*}{circles} & GA & GA (adaptive AB) & -0.106 & 0.064 & -1.670 & 0.095 & [-0.231, 0.018] \\
         & EP (50\%) & EP (adapted) & -0.062 & 0.102 & -0.609 & 0.543 & [-0.263, 0.138] \\
         & GA & EP (50\%) & -0.964 & 0.074 & -13.095 & 0.000 & [-1.109, -0.820] \\
         & GA & Backpropagation & 1.030 & 0.071 & 14.595 & 0.000 & [0.892, 1.168] \\
        \bottomrule
    \end{tabular}} 
    \caption{Performance evaluation of the GA against edge-popup on shown datasets, using the respective sparsity levels that were achieved with our GA configurations as new values for the fixed pruning rates. Depending on the achieved mean accuracy, we either adapt the mean sparsity levels from ``GA'' or from ``GA (adaptive AB)'', which is indicated by the different colored dots. For comparison, we plot the mean accuracies and 95\% confidence intervals for the corresponding GA configuration, backpropagation, and the original edge-popup variant using the default prune-rate of $0.5$. A final statistical analysis evaluates the performance difference of combinations of algorithms based on p-values for the GA and edge-popup configurations, as well as the backpropagation baseline. 
    \subref{tab:statistics} shows the performance deviation of the target algorithm from the reference.
    }
    \label{fig:edge_perform_adapted}
\end{figure*}
Considering the contrary development of the accuracy and sparsity levels at the beginning of the evolution (cf. Fig.~\ref{fig:spars_dev}), we hypothesize a correlation. This also implies a potential connection between the number of pruned parameters throughout the evolution and the final achieved fitness. 
Opposed to us, edge-popup works with fixed pruning rates. To rule out any performance deficits that might arise because of this inflexibility, we include additional edge-popup runs in our comparison study, where we set the pruning rates to the mean sparsity levels that can be achieved with the two GA configurations. We chose that configuration for every architecture and dataset, which scored the highest mean accuracy, and reran the edge-popup experiments with the derived mean sparsity levels.
\\[2pt]
The results of our comparison study are depicted in Fig.~\ref{fig:edge_perform_adapted}. For an extensive evaluation, we plotted the mean accuracy of the best-performing GA configuration for the respective architecture, together with the mean accuracies of backpropagation and the original edge-popup runs. The shaded area around the line plots represents the $95\%$ confidence intervals for the estimation of the mean. Relevant for the comparison of edge-popup with the adapted pruning rates, we specified the respective mean sparsity levels our GA configurations achieved on the x-axis in addition to the architectures. We can see in Fig.~\ref{fig:edge_moons_adapted} that for \texttt{moons}, these levels dropped with increasing network sizes, converging to $0.5$, which corresponds to edge-popup's default prune$\_$rate value. This suggests that the influence of the varied pruning rate should be higher for smaller architectures. 
Indeed, we observe the biggest relative change for network A. The varied prune rate appears to have a negative effect as it resulted in multiple low-accuracy runs, which negatively influenced the mean. Yet, because of the high variance, there are also some instances that scored higher compared to \textit{EP (50\%}). For the other networks, there was little to no change regarding the mean accuracy, and if there was, it was only negative. 
The same holds true for the \texttt{circles} dataset, as can be seen in Fig.~\ref{fig:edge_circles_adapted} except for architecture C. Since there is considerable variance between runs that use the same pruning rate and the confidence intervals mostly overlap, it cannot be concluded with certainty that these changes are due to the varied pruning rates. Overall, none of the changes lead to a significant performance improvement. 
\\[2pt]
If we compare edge-popup against the GA configurations, it becomes apparent that the GA outperforms for every dataset and architecture, even if we enable edge-popup to find sparser subnetworks. In fact, the adapted pruning rates lead to a worse performance. Based on this, we can conclude that the GA can find higher accuracy scoring subnetworks that are also sparser and approximately match backpropagation for larger networks. 
To test the statistical significance of our findings, we fit a linear mixed model to our accuracy data. We are mainly interested in comparing the different algorithms across different architectures on the same dataset. That's why we model the algorithms as fixed effects and treat the four architectures and varying network initializations as random effects to account for the variability across runs. We perform our statistical analysis using the \textit{MixedLM} module from \citep{seabold2010statsmodels}. 
To fit the data and ensure proper convergence, we employ \textit{Powell's algorithm}, use the restricted maximum likelihood (REML), and standardize the accuracies. The results of our analysis are listed in Table \ref{tab:statistics}. 
\\[2pt]
For assessing the statistical significance, we consider various statistics, including coefficients and p-values, to determine the relationship between the reference algorithm and the target algorithm. Starting with the \texttt{moons} dataset, we can see that the coefficient for \textit{GA (adaptive AB)} is positive. This indicates that it performs slightly better than the GA, considering all architectures and initializations. Yet, because the p-value is $> 0.05$, this performance difference is statistically insignificant. The same holds for edge-popup with the varied prune$\_$rate, although in this case, the negative coefficient indicates a slightly worse performance of \textit{EP (adapted)}, supporting our previous findings. 
Because both \textit{GA (adaptive AB)} and \textit{EP (adapted)}'s performances deviate insignificantly from the reference algorithms, we only compare \textit{GA} and \textit{EP (50\%)} against each other. Doing so, we observe a large negative coefficient, implying a considerably worse performance of \textit{EP (50\%)}. This result is statistically significant, as the p-value is 0. Compared to backpropagation, the GA configuration performs moderately worse, which is also a statistically significant result. For the \texttt{circles} dataset, the analysis draws a very similar picture. Although, in this case, \textit{GA (adaptive AB)} has a negative coefficient, supporting the (almost statically significant) result that the base GA configuration is a more appropriate choice for this dataset. Accounting for all random effects, backpropagation here clearly outperforms the GA. 
\\[2pt]
We conclude that the GA performs significantly better than edge-popup in the given scenarios and performs only moderately worse than backpropagation on the \texttt{moons} dataset regarding the final accuracy.


\begin{figure*}[t]\centering
  \subfloat[\texttt{digits}, R=25 \label{fig:GA_fewer_digits}]{\includegraphics[width=0.33\textwidth]{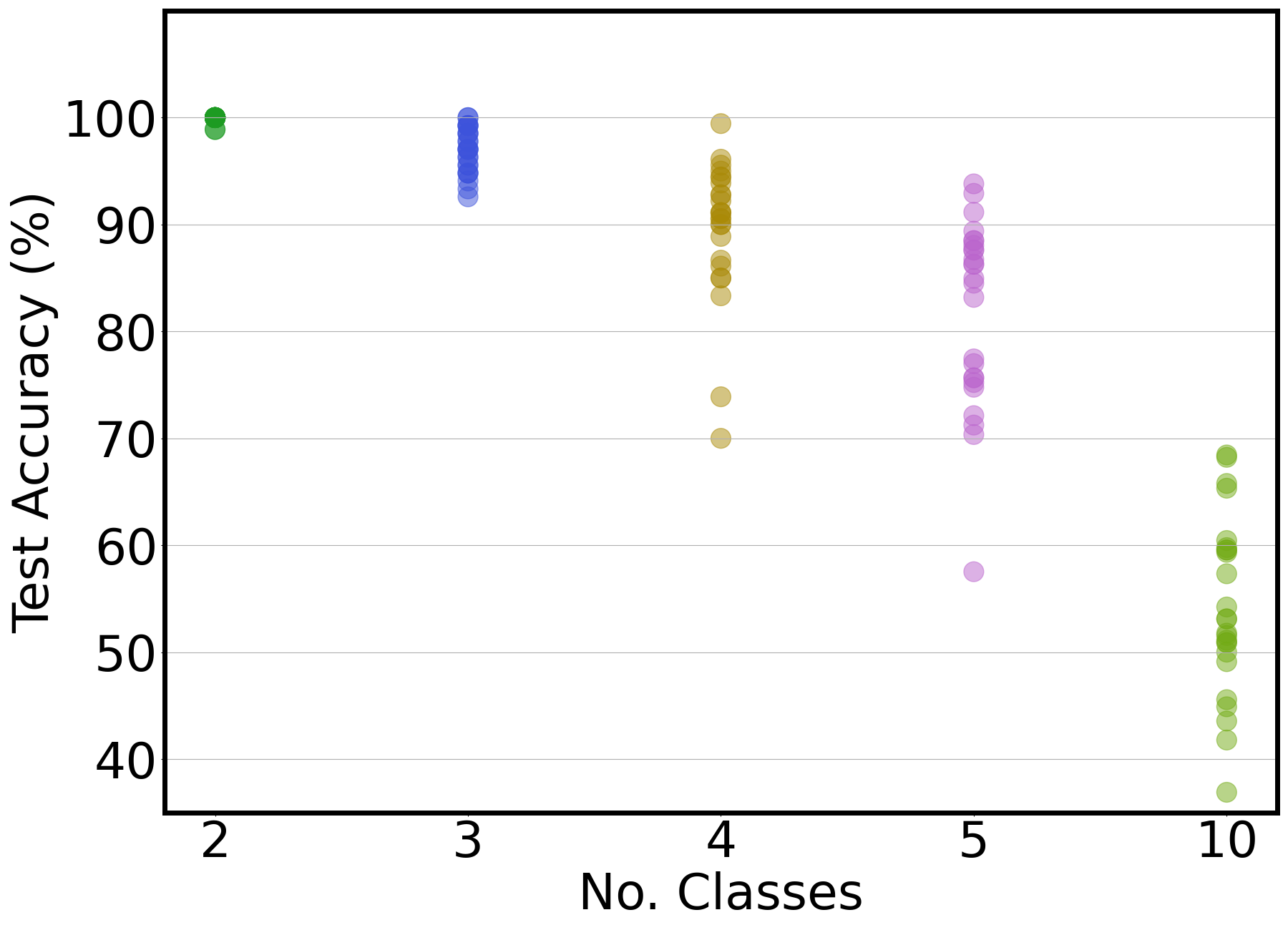}}
  \subfloat[\texttt{blobs} test dataset\label{fig:blobs_dataset}]{\includegraphics[width=0.33\linewidth]{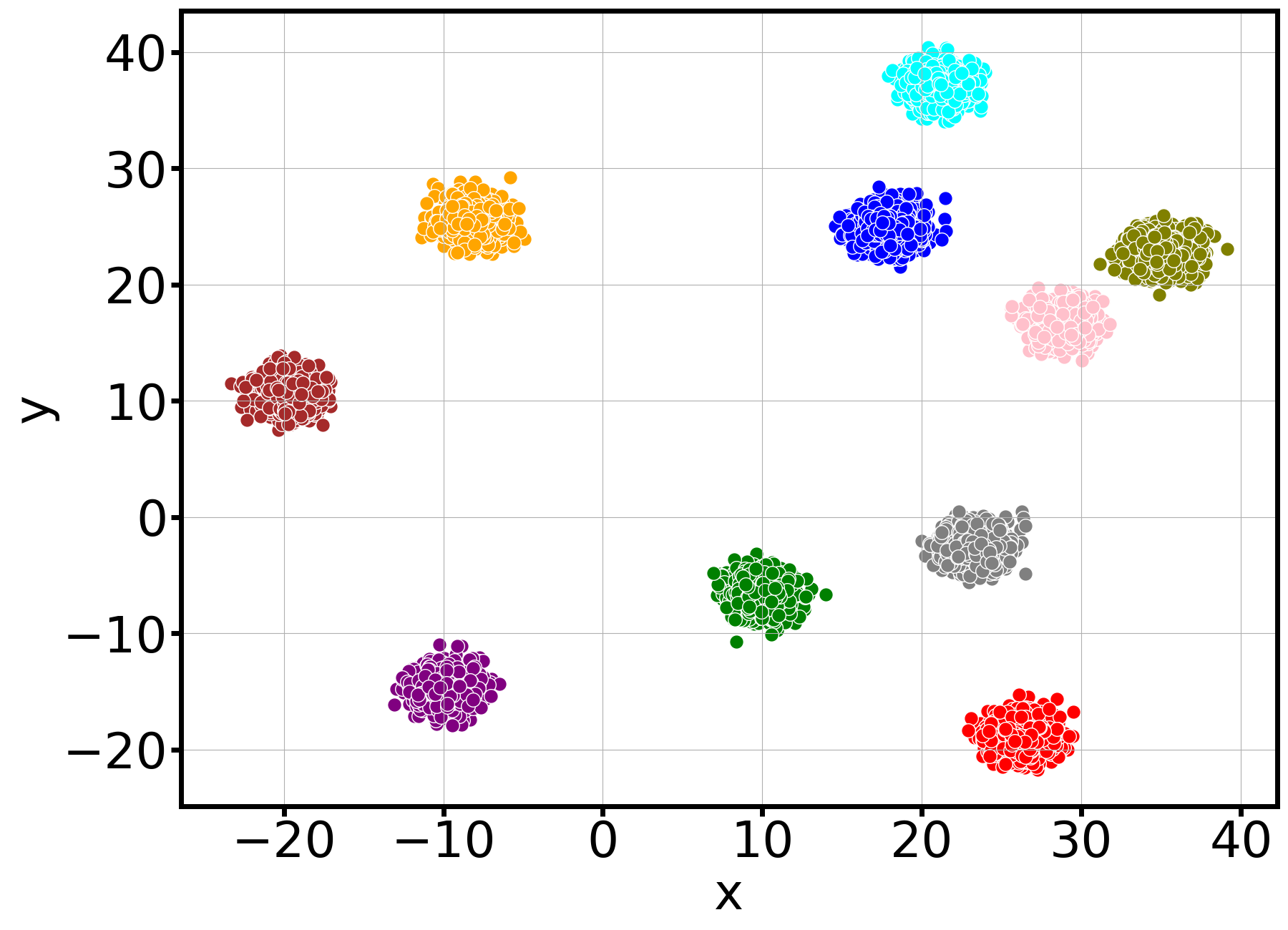}}
  \subfloat[\texttt{blobs} \label{fig:GA_blobs}]{\includegraphics[width=0.33\textwidth]{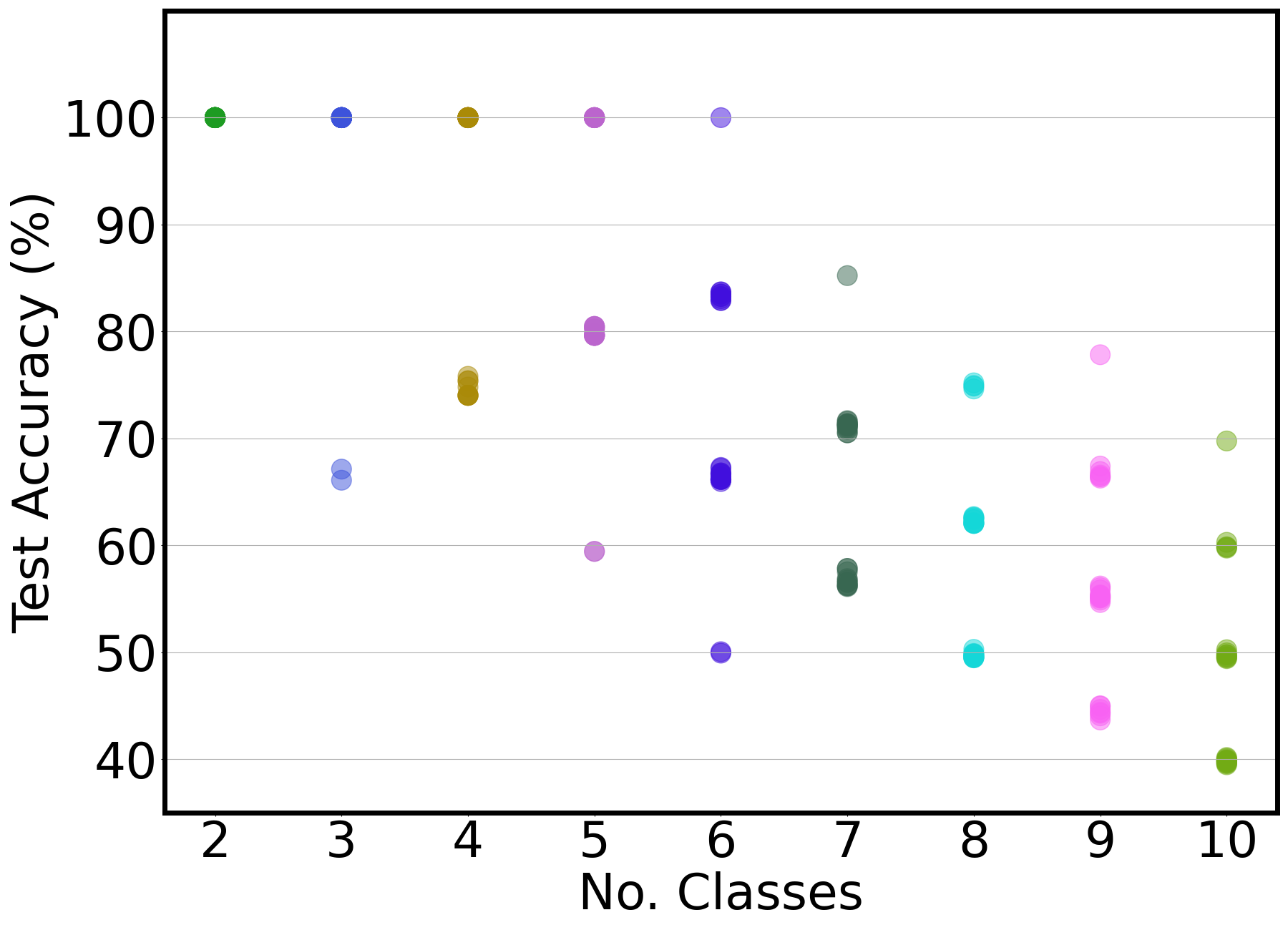}}
  \caption{Mutli-Class performance on $R$ runs: \subref{fig:GA_fewer_digits} Overview of the distribution of final accuracies achieved by the GA (without accuracy bound) for different variants of the \texttt{digits} dataset. For the binary case we consider class labels 0 and 1, for the ternary case 0, 1, and 2, and so on. All runs were performed on architecture B (cf. Table~\ref{tab:architectures}), but we adapted the number of output neurons according to the number of class labels. \subref{fig:blobs_dataset} illustrates the \texttt{blobs} test dataset for $10$ differently labeled cluster of 2d points, generated using \texttt{scikit-learn} \citep{pedregosa2011scikit}. \subref{fig:GA_blobs} Performance of default GA configuration without accuracy bound on the \texttt{blobs} dataset with varying class labels. The dataset was generated using the \texttt{scikit-learn} function \texttt{make\_blobs} and consists of differently labeled clusters, each containing 1250 datapoints. For this experiment, we used a network architecture $[2, 100, n]$ where $n$ is the number of class labels.}
\end{figure*}

\subsection{Multi-Class Performance}\label{subsec:prob_multiclass}

So far, we only considered datasets for binary classification. It turns out that \approach{} has a much harder time finding suitable lottery tickets for multi-class classification problems.
We first analyze that behavior by comparing the performance of the GA using network architecture B\footnote{Preliminary experiments showed the highest \textit{GA} accuracies on this architecture in the computationally less intensive base configuration.} 
and the 2-, 3-, 4-, 5-, and 10-class variants of the \texttt{digits} dataset. 
\\[2pt]
The results are depicted in Fig.~\ref{fig:GA_fewer_digits}. We observe that, at least in the binary case, the GA still reaches perfect accuracy in most of the runs; however, using just one more class label leads to a considerable increase in variance. There are still runs that reach approximately 100\% accuracy, which is not the case anymore for the 4-class and 5-class settings, where the variance further increases, and there is a noticeable drop in achieved accuracy. When we reach the 10-class setting, the mean accuracy is only a little above 54\%. The increasing number of class labels seems to pose a considerable challenge to the GA. 
\\[2pt]
These observations also hold for much simpler multi-class problems: For a follow-up experiment, we introduce the \texttt{blobs} dataset consisting of up to 10 different 2-dimensional Gaussian-shaped clusters with different class labels $1,...,10$. These clusters are uniformly distributed in the feature space and do not overlap, as shown in Figure~\ref{fig:blobs_dataset}. For this experiment, we used a neural network architecture that consists of 2 input neurons, 100 hidden units, and as many output neurons as required, given the number of classes. For the classification of points in 2d space, backpropagation is able to reach 100\% accuracy regardless of the number of classes. 
The results are shown in Fig.~\ref{fig:GA_blobs} and draw a similar picture as the first experiment: While instances with fewer classes can reach perfect accuracy, trying to distinguish more class labels leads to increasingly bad final accuracies. However, in contrast to the \texttt{digits} dataset, the GA can find high-accuracy subnetworks for a higher maximum number of class labels (up to 6 classes), which suggests that the GA can indeed deal with more classes when the input space is less complex.
\\[2pt]
Aside from that, we observe a unique multi-modal distribution of the accuracies, whose detailed analysis is left for future work. At the moment, we reckon that since the type of training we perform with our GA is at its core just the task of solving a complex combinatorial problem, i.e., the problem of sampling proper decision boundaries, the complexity of this task grows superlinearly with an increasing number of decision boundaries that need to be arranged in the feature space. One observation we made during the GA runs is that the accuracy is improved in only very small steps, and the GA takes a long time to converge. This behavior could partly be explained by the low population diversity that leads to very homogenous populations already early in the evolution. In that phase, the main driver of change is the mutation operation, which can only lead to small accuracy improvements. We hypothesize that it needs more sophisticated GA operations, including proper diversity retention techniques, to deal with complex multi-class datasets.

\section{Conclusion}
\label{sec:conclusion}

We have presented a GA-based approach for finding strong lottery ticket networks without any training steps on the network. We have analyzed different configurations and behaviors of the GA and have shown that, for simple binary classification problems, our approach outperforms the start-of-the-art method edge-popup by producing smaller and more accurate subnetworks. This holds even when the latter is given a more beneficial weight initialization procedure. Furthermore, we found that forcing edge-popup to produce subnetworks that possess the same sparsity levels as the ones produced by the GA leads to a drop in accuracy. Although integrating an adaptive accuracy bound resulted in slightly better accuracies on the \texttt{moons} dataset, in our experiments, this effect is statistically insignificant and comes with reduced computational efficiency, favoring the standard GA. Finally, we have also observed that the performance of our approach breaks down when finding networks for multi-class classification problems. This poses substantial questions about the relationship between network structure and learnability for future research.
\\[2pt]
Notably, in the shown example datasets, our GA-based approach has the advantage over edge-popup, which implements training steps via backpropagation and thus depends on gradient information, which our approach does not. This can be seen as a call to revisit alternative methods of evolving neural networks, at least for special cases. Since our approach effectively frames the problem of finding a good neural network as a problem of binary combinatorial optimization, it may also open up new solving methods to this application (see \citet{whitaker2022quantum}) or allow for better integration of neural networks in scenarios where combinatorial optimization is already employed. 
\\[2pt]
We would also like to point out that --- since the GA is not using gradient information --- it is likely that our approach has applications beyond classical neural networks, which are built on functions that allow gradient information to pass through. We hypothesize that using our GA, it should be possible to use non-differentiable evaluation functions like the \textit{edit distance} \citep{levenshtein1966binary} for strings or \textit{logical consistency} checks for propositional logic directly as loss functions without requiring a potentially suboptimal differentiable surrogate (cf. \citet{patel2021feds, li2019augmenting}) which would have important implications for fields like natural language processing or neural reasoning. To allow for the comparison to the state of the art, we chose classification problems for this paper; however, future work should aim for more complex network structures that allow for non-differentiable functions and test if our approach --- and thus a variant of the lottery ticket hypothesis -- still functions there.

\section*{\uppercase{Acknowledgements}}
This work was partially funded by the Bavarian Ministry for Economic Affairs, Regional Development and Energy as part of a project to support the thematic development of the Institute for Cognitive Systems.


\bibliographystyle{apalike}
{\small
\bibliography{GALA}}

\end{document}